\pdfoutput=1
\documentclass[11pt]{article}

\usepackage[]{emnlp2021}

\usepackage{amsfonts}
\usepackage{times}
\usepackage{latexsym}
\usepackage{diagbox}

\usepackage{array}
\usepackage{tabularx}
\usepackage{algorithm}
\usepackage[noend]{algpseudocode}
\usepackage{xparse}

\makeatletter
\NewDocumentCommand{\LeftComment}{s m}{%
  \Statex \IfBooleanF{#1}{\hspace*{\ALG@thistlm}}\(\triangleright\) \textit{#2}}
\makeatother

\newlength\myindent
\newcommand{\minus}{\scalebox{0.75}[1.0]{$-.$}}
\newcommand{\plus}{\scalebox{0.75}[1.0]{$+.$}}

\algnewcommand{\parState}[1]{\State%
  \parbox[t]{\dimexpr\linewidth-\algmargin}{\strut #1\strut}}
\usepackage{microtype}
\usepackage{amsmath}
 \usepackage{multirow}
\usepackage[position=bottom]{subfig}
\usepackage{booktabs}
\usepackage{CJKutf8}
\def\aclpaperid{473}
\usepackage{xspace}
\usepackage{marvosym}

\usepackage{pgf}
\usepackage{collcell}
\usepackage{booktabs}
\usepackage{tcolorbox}

\newcommand*{\MinNumber}{-0.5}%
\newcommand*{\MaxNumber}{0.5}%
\newcommand{\ApplyGradient}[1]{%
  \pgfmathsetmacro{\PercentColor}{100.0*((#1+\MaxNumber)/(\MaxNumber-\MinNumber))}%
  \edef\x{\noexpand\cellcolor{red!\PercentColor}}\x\textcolor{white}{#1}%
}
\newcolumntype{R}{>{\collectcell\ApplyGradient}{r}<{\endcollectcell}}
\newcommand{\eg}{\textit{e.g.,}\xspace}
\newcommand{\ie}{\textit{i.e.,}\xspace}
\newcommand{\Fone}{F\textsubscript{1}\xspace}
\newcommand{\pop}{\textbf{+P}\xspace}
\newcommand{\adv}{\textbf{+A}\xspace}
\newcommand{\tac}{TAC\xspace}
\newcommand{\wiki}{Wiki\xspace}
\newcommand{\mention}[1]{\textit{#1}}
\newcommand{\kb}[1]{\textsl{#1}}

\title{Improving Zero-Shot Multi-Lingual Entity Linking}

\author{Elliot Schumacher \hspace{1em}
        James Mayfield \hspace{1em}
        Mark Dredze \hspace{1em} \\ 
        Johns Hopkins University \\
        \texttt{\{elliotschumacher,mayfield,mdredze\}@jhu.edu}}

\date{}
\begin{document}

\maketitle

\begin{abstract}
Entity linking -- the task of identifying references in free text to relevant knowledge base representations -- often focuses on single languages.  We consider multilingual entity linking, where a single model is trained to link references to same-language knowledge bases in several languages.  We propose a neural ranker architecture, which leverages multilingual transformer representations of text to be easily applied to a multilingual setting. We then explore how a neural ranker trained in one language (\eg English) transfers to an unseen language (\eg Chinese), and find that while there is a consistent but not large drop in performance.  How can this drop in performance be alleviated? We explore adding an adversarial objective to force our model to learn language-invariant representations.  We find that using this approach improves recall in several datasets, often matching the in-language performance, thus alleviating some of the performance loss occurring from zero-shot transfer.

\end{abstract}

\section{Introduction}

\begin{figure*}[!htb]
    \centering
    \begin{minipage}{.48\textwidth}
        \centering
    \begin{tcolorbox}[width=0.98\textwidth,
                  boxsep=0pt,
                  arc=0pt,
                  colback=white,
                  left=2pt,
                  right=2pt,
                  top=2pt,
                  bottom=2pt
                  ]%%
                 También lo acompañan el presidente del \textbf{\mention{Senado}}, Luis Miguel Barbosa y los
industriales Valentín Diez Morodo, ...
    \end{tcolorbox}
                \centering
\begin{tabularx}{0.9\linewidth}{@{}r|X@{}}
name & \kb{Senado de la República} \\ \midrule
desc. & \kb{El Senado de los Estados Unidos Mexicanos es la Cámara Alta  ...} \\ \midrule
type & \kb{governmental\_body}
\end{tabularx} 
    \end{minipage}%
    \begin{minipage}{0.48\textwidth}
        \centering
         \begin{tcolorbox}[width=0.98\textwidth,
                  boxsep=0pt,
                  arc=0pt,
                  colback=white,
                  left=2pt,
                  right=2pt,
                  top=2pt,
                  bottom=2pt
                  ]%%
                As the general explained in
testimony to the \textbf{\mention{Senate}} Armed Services Committee, “The enduring hostilities...
    \end{tcolorbox}
                \centering
\begin{tabularx}{0.9\linewidth}{@{}r|X@{}}
name & \kb{United States Senate} \\ \midrule
desc. & \kb{The United States Senate is the upper chamber of the United ...} \\ \midrule
type & \kb{governmental\_body}
\end{tabularx}

    \end{minipage}
\caption{Example Spanish mention \mention{Senado}, which is a link to the Spanish KB entity \kb{Senado de la República} (the Senate of Mexico), and the example English mention \mention{Senate}, referring to the entity \kb{United States Senate}.}
\label{fig:ex_arch}
\end{figure*}

%Overview of mlel
Entity linking -- the process of matching mentions of people, places or organizations with a relevant knowledge base (KB) entry -- has often focused on English text and KBs.  This focus omits non-English knowledge bases, which often have better setting-specific details available compared to the English version.  Linking non-English text to an English knowledge base -- referred to as Cross-Lingual Entity Linking -- also is not appropriate for non-English speakers.  However, relying on non-English entity linking annotations alone is limiting, as there is far less supervised data available in some languages, such as Spanish, than in English.  Therefore, training a Multi-lingual entity linking model -- building a single model for several pairs of source text and KB languages -- is an important area of research.

%Walkthrough of example
Consider the example in Figure \ref{fig:ex_arch}, of the Spanish language mention \mention{Senado} (\mention{Senate}), a link to the KB entry \kb{Senado de la República} (\kb{Senate of the Republic of Mexico}).  An entity linking model uses the information available in the text, such as the mention itself and the surrounding context, and in the knowledge base, such as the name, description, and type(s), to score the likelihood of a match.  Many approaches to entity linking learn these linkages by training on a set of Human-annotated links in the target language.  If there are no or few language-specific annotations available, how can we train a model on a annotation-rich language like English which will perform well on non-English languages?

%Other approaches

%General linker approach
We propose a neural approach to multi-lingual entity linking.  We use a cross-lingual pretrained transformer model, XLM-Roberta (XLM-R) \cite{conneau2019unsupervised}, to build representations of the available text for a given mention and candidate entity pair -- span text and surrounding context for the mention, and name and description for the entity.  We then feed each of these representations through a feed forward neural model to produce a likelihood score.  As XLM-R has been trained cross-lingually, it has been shown to build robust representations of text in a wide variety of languages. However,  how does a model perform if it has been trained on in-language annotated data compared to one without such annotations?  We find that even with the cross-lingual ability of XLM-R, in-language annotation data is key to an accurate multi-lingual linking model.

%Adversarial and pop objectives
We propose two methods to improve the performance of zero-shot multilingual entity linking.  First, we leverage the data available in the target language KB to capture the likely popularity of each entity.  Second, we add an adversarial objective when training the linker, adapting a method proposed in \citet{chen-cardie-2018-multinomial}.  For this objective, we add a weight-tied hidden layer as an intermediate step between the name and mention representation, and the rest of the linker. In addition to using the output in the ranker, we also use the respective name and mention output to train a language classifier.

We then use a two-step training procedure during each training iteration.  First, we train the language classifier alone to predict the incorrect language label for unannotated portions of the source (e.g. English) and target (e.g. Spanish) text.  Second, we jointly train the ranker and the language classifier using the correct source (e.g. English) language labels.  This is designed to force the representation of both the name and mention to be language-independent.  We find that both model adjustments improve zero-shot performance on several language pairs, and that the adversarial model specifically produces consistent improvement in recall.

\section{Entity Linking Model} \label{sec:model}

\begin{figure*}[!htb]
    \centering
    \begin{minipage}{.5\textwidth}
    \centering
        \includegraphics[width=0.8\textwidth]{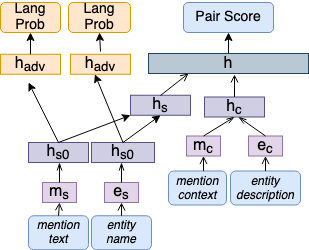}

    \end{minipage}%
    \begin{minipage}{0.3\textwidth}
    \centering
        \includegraphics[width=0.65\textwidth]{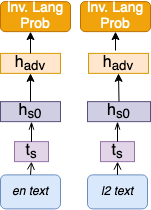}

    \end{minipage}
\caption{Example Spanish mention \textit{Senado}, which is a link to the Spanish KB entity \textit{Senado de la República} (the Senate of Mexico), and the example English mention \textit{Senate}, referring to the entity \textit{United States Senate}.}
\label{fig:arch_adv}
\end{figure*}

In Figure \ref{fig:ex_arch}, there are two example mentions shown -- one in Spanish (\mention{Senado}) and one in English (\mention{Senate}). In each case, the mentions are linked to entities in the KB in the same language -- \kb{Senado de la República} for the Spanish mention and \kb{United States Senate} for the English mention.  In addition to the text of the mention and the name of the entity, there are other sources of information available in the context of the mention (the surrounding sentences) and the entity (the description), and the mention and entity types.

One approach to entity linking would be to train single-language models for each case.  However, this approach has several limitations.  First, while some languages such as English have a large amount of entity linking annotated data, others, such as Spanish, have far less.  Second, training single-language models requires retraining for each new language, which does not scale well.  On the other hand, a multilingual entity linking approach trains a single model that can be applied to several languages.  This allows for sharing information between languages -- leveraging English training data for Spanish linking, for example.

\subsection{Architecture}\label{sec:arch}

To score a mention $m$ and and entity $e$, we leverage a pointwise neural ranker inspired by the architecture discussed in~\citet{dehghani2017neural}.  This produces a score for each mention-entity pair, thus creating a ranking of entities specific to each mention.  Additionally, this pointwise approach allows for scoring of previously unseen entities.  We select the entities to score for each mention by using a Triage system, as discussed in Section \ref{sec:dataset}.

We designed our ranker to capture two common sources of information about the entity -- the mention string and entity name, and the context of the mention and the entity description.  We explored adding type information, but found that did not improve performance significantly.  We create separate multilingual representations for the mention string and entity name ($m_s$ and $e_s$), and the mention and entity context ($m_c$ and $e_c$).  The string and context pairs are fed into separate multilayer perceptrons (MLP), outputting an embedding that models the relationship between the entity and the mention.  For example, we input $m_s$ and $e_s$ into a text-specific hidden layer which outputs a combined representation $r_s$, and we input $m_c$ and $e_c$ into a context-specific hidden layer which outputs a representation $r_c$.  These representations $r_s$ and $r_c$ are then fed into a final MLP, which produces a score between $-1$ and $1$. 

To train our model parameters $\theta$, we score a mention $m$ and a correct entity link $e_{+}$, and separately score the same mention paired with $n$ randomly sampled negative entities $e_{-}$.  We apply hinge loss between the positive pair and the best performing negative pair; 
\begin{multline*}
    \mathit{L}(\theta) = \mathbf{max}\{0,\epsilon-(S(\{m,e_{+}\};\theta)-\\ \mathbf{max}\{S(\{m,e_{0-}\};\theta) \dots 
            S(\{m,c_{n-}\};\theta)\}\}
\end{multline*}
We use the resulting loss to back propagate through the entire network.  We use random combinations of parameters to select the best model configuration, which is discussed further in the Appendix.

\subsection{Multilingual Representations}\label{sec:rep}

%The mention $m$ and entity $e$ both have a name and context/description. While the entity is in English, the mention may be in a different language. We seek a multilingual representation to allow comparisons across languages.

%We seek to create multilingual representations for both $m$ and $e$.  Entity linking literature focuses on three features - text, context, and type - which previous research has shown result in strong performance~\cite{ratinov-etal-2011-local, francis2016capturing, upadhyay-etal-2018-joint}; we create representations for each of these.

To create representations of the name and context for a mention-entity pair, we use XLM-Roberta (XLM-R)~\cite{conneau2019unsupervised}, a multilingually-trained transformer representation model.  XLM-R has been shown to outperform other transformer models (such as mBERT~\cite{devlin-etal-2019-bert}) at multilingual tasks, and we found the same performance pattern in early linker experiments.
Consider the Spanish example in Figure~\ref{fig:ex_arch}. First, we create a representation of the mention text $m_s$, \mention{Senado}, by feeding the entire sentence through XLM-R, and form a single representation using max pooling on only the subwords of the mention. We create a representation of the entity name $e_s$, \kb{Senado de la República} in the same way, except without any surrounding context.

To create $m_c$, we select the sentences surrounding the mention, up to XLM-R's sub-word limit.  The resulting text is fed into XLM-R, and we again use max pooling to create a single representation.  A similar method is used for the entity context $e_c$ , but uses the definition or other text in the KB, using the first 512 subword tokens from that description. 

\section{Multilingual Transfer}
Our linking model is inherently multilingual due to the representation power of XLM-R, and therefore a single model can be trained on several languages.  However, we find that while applying a multilingual model trained on a single language (e.g. English) to a previously unseen language (e.g. Spanish) results in reasonable performance, there is a significant performance drop compared to a in-language trained model.  

We propose two methods of improving this zero-shot performance.  One source of error can stem from the model scoring mentions and entities in an unseen language. Considering the example in Figure \ref{fig:ex_arch}, how well does the model recognize that the English mention \mention{Senate} and the Spanish mention \mention{Senado} should be treated similarly? Therefore, we explore adding an adversarial objective to ensure that the model is forced to learn language-agnostic representations of the text, which will better transfer to other languages.  This does not require annotated mention text, but rather can work with unstructured text in the target language.  

Another source of error comes from the model scoring entities is has not seen in any language. Again considering the example in Figure \ref{fig:ex_arch}, how well does the ranker recognize that the Spanish-language mention \mention{Senado} may be more likely to refer to the Mexican Senate, as opposed to the U.S. Senate, compared to English-language text? For this issue, we explore the importance of adapting the model to the specifics of the knowledge base, by adding popularity measures drawn from the KB into our model.  While this does rely on KB-specific information, annotated mentions are not required. Separately, both of these adaptations improve the performance of transferred multilingual entity linking models, and together regain the largest amount of performance loss compared to in-language models.

\subsection{Language Adaptation}
\begin{algorithm}[]
\begin{algorithmic}[1]
\Require
Mentions $\mathbb{M}$, entity labels $\mathbb{E}$ and labels $\mathbb{N}$; English Text $\mathbb{A}$ and labels $\mathbb{B}$;  Unlabeled L2 Text $\mathbb{C}$ and labels $\mathbb{D}$; Hyperpamameter $\lambda > 0$, $y$, $z \in N$
\Repeat
\State $l_{adv}$, $l$ = 0
\For {$i = 0$ to $y$}
\Comment{Adversarial Step}
\State $t_{A}$ = representation of $\mathbb{A}_{i}$
\State $t_{C}$ = representation of $\mathbb{C}_{i}$

\State $p_{A}$ = $h_{adv}(h_{s0}(t_{A}))$
\State $p_{C}$ = $h_{adv}(h_{s0}(t_{B}))$
\LeftComment{Calculate Lang scores}

\State $l_{adv}$ += MSE($p_{A}$, $\mathbb{D}_{i}$)
\State $l_{adv}$ += MSE($p_{C}$, $\mathbb{B}_{i}$)
\LeftComment{Calculate Loss using reversed labels}

\EndFor
\State Update $h_{adv}$ using $l_{adv}$

\For {$i = 0$ to $z$}
\Comment{Main Step}
\State $m$ = representation of $\mathbb{M}_{i}$
\State $r_{m}$ = $h_{s0}(m)$
\State $e$ = representation of $\mathbb{E}_{i}$
\State $r_{e}$ = $h_{s0}(e)$

\State $l$ = Entity-Mention Loss (Eq. 1) with \hspace*{15mm} $r_{m}$ and $r_{e}$ 

\State $p_{M}$ = $h_{adv}(r_{m})$
\State $p_{E}$ = $h_{adv}(r_{e})$

\LeftComment{Calculate Lang scores}

\State $l$ += $\lambda$ (MSE($p_{M}$, $ \mathbb{N}_{i}$) + MSE($p_{E}$, $ \mathbb{N}_{i}$))
\LeftComment{Calculate Loss using correct labels}

\EndFor
\State Update all parameters except $h_{adv}$ using $l$

\Until{convergence}

\end{algorithmic}
\caption{Pseudo-code of model training using the adversarial objective.  First, a small set of randomly selected text (in our experiments, $y=5$) is used to adversarially train the language classifier.  Second, the main entity linking loss and the language classifier, with the correct labels, are jointly trained.  }
\label{alg:adv}

\end{algorithm}

To force English-trained models to learn representations better suited for transferring to unseen languages, we look to \citet{chen-cardie-2018-multinomial}.  The authors build a text classification system which uses an adversarial objective that forces the network to learn domain-invariant features. In addition to their standard text classifier which takes as input features from a shared and domain specific feature extractor, they add a domain discriminator which uses the shared feature extractor as input.  They run two training passes -- a normal training pass for the entire network, which uses the correct classification and domain labels.  Second, they adversarially train the domain discriminator and the shared feature extractor only, using the inverse of domain labels as the target.  During prediction, the domain discriminator portion of the network is ignored.  They find this objective improves performance when classifying text from previously unseen domains.

Instead of forcing our model to learn domain-invariant representations, we force it to learn language-invariant representations.  As our models are all trained on a single language, it is likely the linker learns language-specific patterns.  We want to force the linker to learn patterns that can apply to any language input, and to do so without relying on annotated linking data in a second language.
As illustrated in Figure \ref{fig:arch_adv}, we add a weight-tied hidden layer, $h_{s0}$, which takes as input the representations of the mention text $m_s$ and entity name $e_s$.  The resulting representation is then input to the linker, and also to a feed-forward neural network $h_{adv}$.  This language classifier outputs the likelihood of the text being a specific language.  We then change our training iteration for a specific batch in the following way.  

As detailed in Algorithm \ref{alg:adv}, we run an adversarial pass, which first creates representations of English $\mathbb{A}$ and L2 $\mathbb{C}$ text, using the same method as for $m_s$. Each of the two representations are fed into the shared invariant layer $h_{s0}$, the language classifier $h_{adv}$, and softmaxed to produce separate language likelihood scores forthe  English $p_{A}$ and L2 $p_{C}$ text. Importantly, we calculate the mean squared error (MSE) using the inverted language labels -- for the English input, we apply the L2 label $\mathbb{D}$, and for the L2 input, we apply the English label $\mathbb{B}$. For some settings, we stop training the adversarial step after 50 epochs.

Second, we run a standard training pass, in which we jointly train the linker and the language classifier.  The entity linking loss is unchanged from \S \ref{sec:arch}, except that the $m_s$ and $e_s$ are first fed separately through the shared invariant layer $h_{s0}$.  The loss for language classifier is unchanged from the first step except that the correct labels $\mathbb{N}$ are used.  The effect of the language classifier loss is controlled by the parameter $\lambda$, which we set to be either $0.25$ or $0.01$ depending on the setting.  Models including this are referred to as \adv.  Further implementation details are available in the Appendix. We experimented with simply adding the additional layers $h_s{0}$ and not applying the adversarial objective, and feeding both the language-invariant  (\eg $m$) and language-specific representations (\eg $r_{m}$)) into the linker.  However, both configurations performed worse in early development experiments than the proposed model.

\subsection{KB Adaptation}

When making a linking decision for the mention \mention{Senado}, the linker must consider that some entities (\eg \kb{Senado de la República}, or \kb{Senate of Mexico}) are more likely than others (\eg \kb{Senado de Arizona}, or \kb{Senate of Arizona}).  One source of this information is from annotated entity linking data -- as discussed previously, this is expensive to annotate.

Therefore, we apply a common entity linking approach and leverage the cross-links within the entities in the KB, which are a good indicator of the popularity of entities.  For example, the entity \kb{Senado de la República} might have a link to the lower legislature of Mexico, \kb{Cámara de Diputados}, and the President of Senate, \kb{Presidente de la Cámara de Senadores}.  Others, such as \kb{Senado de Arizona}, are likely to have fewer.  We simply count unique cross-links between entities, divide by the median number of such links, and feed the resulting number into the final feed forward neural network $h$.  Models including this are referred to as \pop.

\section{Datasets}\label{sec:dataset}
We conduct our evaluation on two cross-language entity linking datasets.

{\bf TAC.}
The 2015 TAC KBP Entity Discovery and Linking dataset~\citep{Ji2015} consists of newswire and discussion form posts in English, Spanish, and Mandarin Chinese.  The training set consists of mentions across 447 documents, and the evaluation set consists of mention annotations across 502 documents. A mention is linked to NIL if there is no relevant entity in the KB, and the KB is derived from a version of BaseKB. As many entities in the KB do not have any non-English names available, those are omitted from the dataset.  This leaves us $14,793$ development mentions, of which $11,344$ are non-NIL.
%FINISH NUMBERS

{\bf Wiki.}
We created a multi-lingual entity linking dataset from Wikipedia links~\cite{pan2017cross} that includes Farsi and Russian. A preprocessed version of Wikipedia\footnote{We thank the authors of ~\citet{pan2017cross} for providing us with a preprocessed Wikipedia.  We will work with the authors to release the dataset.} is annotated with links to in-language pages, which we treat as an entity.  We treat each language as having a distinct KB, although some entities may overlap in different languages. We consider this to be silver-standard data because--unlike the \tac dataset--the annotations have not been reviewed by annotators. Some BaseKB entities used in the \tac dataset have Wikipedia links provided; we used those links as seed entities for retrieving mentions, retrieving a sample mention of those and adding the remaining links in the page. We mark 20\% of the mentions as NIL.
%TODO numbers

We predict NILs by applying a threshold; mentions where all entities are below a given threshold are marked as NIL. The threshold is selected based on the respective development set, and is $-1$ unless otherwise noted.  We evaluate all models using 
the evaluation script provided by~\citet{Ji2015}, which reports Precision, Recall, \Fone, and Micro-averaged precision.

\textbf{Triage.}
We use the triage system of~\citet{upadhyay-etal-2018-joint}, which is largely based on work in~\citet{Tsai2016}, allowing us to score a smaller set of entities for each mention, where we assume the mention boundaries are already known.  For a give mention $m$, a triage system will provide a set of $k$ candidate entities ${e_1  \dots e_k}$.   The system uses Wikipedia cross-links (as discussed previously) to generate a prior probability $\mathtt{P}_\mathtt{prior}(e_i|m)$ by estimating counts from those mentions. Originally, this system was designed to produce cross-lingual candidate links, specifically for a non-English mentions to English titles.  We tweak this approach by applying the same pipeline, but for in-language titles, which did not require any major algorithmic adaptations.  Further implementation details specific to the \tac dataset are provided in Appendix B.

\begin{table*}[tb]
    \centering
\begin{tabular}{l|rrrrr|rrrrr}
& \multicolumn{5}{c|}{ es } & \multicolumn{5}{c}{ zh } \\
Model & micro & p & r & \Fone & nn \Fone & micro & p & r & \Fone & nn \Fone \\ \hline
same & 0.623 & 0.910 & 0.711 & 0.798 & 0.870 & 0.670 & 0.862 & 0.787 & 0.822 & 0.844 \\  \hline
nn & \minus248 & \plus015 & \minus078 & \minus047 & \minus060 & \minus426 & \plus048 & \minus067 & \minus019 & \minus018 \\ 
en & \minus057 & \plus015 & \minus076 & \minus045 & \minus059 & \minus299 & \plus032 & \minus140 & \minus072 & \minus087 \\
en+A & \minus007 & \plus013 & \minus005 & \plus002 & \plus006 & \minus198 & \plus015 & \minus017 & \minus002 & \minus006 \\
en+P & \plus009 & \plus010 & \minus095 & \minus061 & \minus079 & \minus208 & \plus008 & \minus150 & \minus089 & \minus110 \\
en+PA & \plus005 & \plus011 & \minus078 & \minus048 & \minus062 & \minus048 & \plus009 & \minus088 & \minus047 & \minus054 \\ \hline \hline
& \multicolumn{5}{c|}{ fa } & \multicolumn{5}{c}{ ru } \\
& micro & p & r & \Fone & nn \Fone & micro & p & r & \Fone & nn \Fone \\ \hline
same & 0.838 & 0.902 & 0.958 & 0.929 & 0.908 & 0.526 & 0.729 & 0.827 & 0.775 & 0.721 \\  \hline
nn & \minus446 & \minus341 & \minus008 & \minus224 & \minus323 & \minus305 & \minus175 & \minus008 & \minus105 & \minus136 \\ 
en & \minus215 & \minus154 & \minus024 & \minus098 & \minus134 & \minus114 & \minus031 & \minus012 & \minus022 & \minus025 \\
en+A & \minus340 & \minus285 & \minus040 & \minus192 & \minus269 & \minus159 & \minus131 & \plus023 & \minus066 & \minus087 \\
en+P & \minus211 & \minus201 & \plus000 & \minus120 & \minus168 & \minus101 & \minus129 & \plus014 & \minus068 & \minus088 \\
en+PA & \minus255 & \minus223 & \minus028 & \minus144 & \minus200 & \minus147 & \minus168 & \plus006 & \minus096 & \minus125 \\ 
\end{tabular} 
        \caption{Compared to an in-language trained model and a nearest-neighbor baseline (\textbf{nn}), how does a zero-shot model trained only on English transfer?  We find that while there is usually a performance improvement, it is often not large.  Can we recover some of that lost performance by using an adversarial objective (\adv) or adding knowledge base information (\pop), or both (\textbf{+PA})?  We find that when applying an adversarial objective specifically, recall is increased leading to higher \Fone scores.  For each setting, we report the change in Micro-avg, precision, recall, \Fone, and non-NIL \Fone on \tac and \wiki datasets compared to the in-language baseline. }
    \label{tab:lang_results}
\end{table*}

\begin{table}[tb]
    \centering
\begin{tabular}{l|ccc}         
    Target & micro & \Fone & nn \Fone \\ \hline
baseline (zh match) & 0.484 & 0.672 & 0.797 \\ \hline
zh adv & \plus009 & \plus014 & \plus015 \\
zh pop & \plus030 & \minus025 & \minus031 \\  \hline \hline
baseline (es match) & 0.472 & 0.678 & 0.802 \\  \hline
es adv & \plus004 & \minus014 & \minus017 \\
es pop & \plus011 & \minus036 & \minus043 \\ 
    \end{tabular}
    \caption{Compared to a baseline English \tac model (with training set size reduced to the noted language's training set size), how does the performance change when a L2 adversarial objective or KB adaptation is applied?  We find that while there are small differences, English performance is largely unchanged.}
    \label{tab:eng_perf}
\end{table}

\begin{table}[tb]
    \centering
\begin{tabular}{cc|cccc}
& Test & micro & r & \Fone & nn \Fone \\ \hline
\multicolumn{2}{c|}{zh} & 0.674 & 0.789 & 0.824 & 0.846 \\ \hline
\multicolumn{2}{c|}{en baseline} & \minus341 & \minus123 & \minus060 & \minus071 \\  \hline

\multirow{4}{*}{\rotatebox[origin=c]{90}{\adv name}} & .25 & \minus190 & \minus001 & \plus009 & \minus003 \\ 
& .01 &\minus202 & \minus078 & \minus033 & \minus036 \\ 
& .25+EL & \minus205 & \minus123 & \minus062 & \minus073 \\ 
& .01+EL & \minus230 & \minus137 & \minus072 & \minus087 \\ \hline
\multirow{4}{*}{\rotatebox[origin=c]{90}{\adv desc}} & .25 & \minus317 & \minus048 & \minus015 & \minus012 \\ 
& .01 & \minus169 & \minus088 & \minus041 & \minus046 \\ 
& .25+EL & \minus287 & \minus188 & \minus108 & \minus133 \\ 
& .01+EL & \minus145 & \minus150 & \minus080 & \minus097 \\ 
\end{tabular} 
    \caption{How do different adversarial settings effect performance?  We consider three variables - the adversarial coefficient $\lambda$, type of text (either names or descriptions), and entity only training for 50 more epochs (\ie we stop updating the language classifier, noted as + EL).  Comparing an in-language trained model to an English trained model using \tac Chinese data evaluation data, we find that while name data performs better than longer text, the difference is slight. Our reported setting of $\lambda = 0.25$ with name data performs best in terms of recall (r), \Fone, and nn \Fone.  However, best performance for micro-average is achieved when using desc. data, $\lambda = 0.01$, and entity linking only training. }
    \label{tab:adv_params}
\end{table}

\section{Model Evaluation} \label{sec:model_eval}
First, we explore how well English trained models transfer to previously unseen languages.  For each language in our datasets -- Spanish (es) and Chinese (zh) for \tac, and Russian (ru) and Farsi (fa) for \wiki, we train a model using in-language training data and compare its performance to a model only trained on the English \tac data.  Note that for English-trained \wiki models, this means that the linker must also efficiently transfer to a new KB. Importantly, we ensure that the training sets in each setting are of equivalent size, by randomly downsizing the larger training dataset (e.g. English) to match the smaller size (e.g. Spanish).  We do this to control for the effect of training set size.  For comparison, we include a simple nearest neighbor baseline (noted as \textbf{nn}).  This model scores each mention-entity pair using the cosine similarity between the mention name representation $m_s$ and the entity representation $e_s$, and selects the highest-scoring pair.

Second, we apply our language (noted as \adv) and KB (noted as \pop) adaptation strategies to each language, and further explore the effect on the model's English performance. In all cases, reported metrics are averaged over three runs. We report results for each language in the form of micro-averaged precision (micro), precision (p), recall (r), and \Fone. We also report separate scores for non-NIL mentions only (nn \Fone).  For development results, see Tables 2 and 3 in the Appendix.

\subsection{Zero-shot performance}

In most languages -- shown in Table \ref{tab:lang_results} -- there is a significant performance reduction when using English training data as opposed to in-language training data.  In the \tac languages, there is a small increase in precision, but a larger decrease in micro-avg, and \Fone.  A similar pattern occurs with \wiki languages (fa and ru), except that precision decreases more significantly than recall. Overall, the drop in performance is not large - the largest drop in \Fone is only $.1$ less than the in-language baseline.  This illustrates that the linker is able to transfer across language and knowledge bases effectively.  Compared to the baseline nearest neighbor model, which one performs better depends on the language.  For example, while Spanish \Fone is nearly the same, Chinese \Fone is slightly higher with the \textbf{nn}, but in Farsi the English-trained model is a large improvement for \Fone.

\subsection{Language and KB adaptation}
Based on experimentation on development, we train the \tac and \wiki datasets with different configurations.  For \tac, we set $\lambda = 0.25$ and apply the adversarial step for the entire training period.  For \wiki, we set $\lambda = 0.01$ and don't apply the adversarial step after 50 epochs. We discuss this further in \S \ref{sec:adv_design}.

When applying the adversarial objective to English-trained models, in most cases we see an increase in recall compared to the baseline English-trained models, and often even compared to the in-language trained models.  For example, the English-trained, Chinese-tested model sees a large drop in recall which is almost completely eliminated when applying the adversarial objective.  This increase in recall leads to nearly-equivalent \Fone performance in Spanish and Chinese in-language models and English trained models with the adversarial objective.  Applying the adversarial objective appears to improve recall, which is useful for settings where high-recall is desired, such as in search. The exception to this is Farsi -- this is likely because the high recall $0.934$ of the zero-shot model already has strong performance.  Comparing to the nearest neighbor baseline, the \adv outperforms the baseline in all languages for \Fone, nn \Fone, micro-avg., and recall.

The addition of KB information appears to have the largest effect on micro-average precision, allowing the model to better perform on rarer entities.  While in Chinese the drop in micro-average is larger in the \pop models than in \adv, in all other cases the micro-average better improves over the English-trained baseline.  Combining the two approaches has mixed effects.  In Chinese, the resulting increase in recall is smaller than the adversarial increase over the baseline, but the micro-average performs much better.  In Spanish, the recall and micro-average are essentially unchanged from the baseline.

We also conducted a type-level analysis of performance on the \tac evaluation dataset using provided mention types (see Appendix Table 2).  When comparing the performance between in-language, English, and English\adv trained models, the performance of \kb{Person} mentions are consistent.  This would be expected in Spanish but is also true in Chinese, which does not share a script with the training language.  The largest performance change occurred in \kb{Geo-Political Entities}.  For Chinese, \Fone drops $0.15$ for an English trained model compared to an in-language trained model, but the deficit is erased in the English\adv model.  A similar pattern occurs in Spanish, suggesting that the adversarial model is able to improve the more challenging entity types.

\subsection{Effect on English performance}

What effect does forcing an English-trained model to be better suited to a second language have on English-language performance?  In Table \ref{tab:eng_perf}, we report results on \tac English evaluation data in three settings.  First, a baseline linker trained on English training data matched to the size of the target language's training data.  Second, we apply the adversarial object, and third, the KB adaptation.  Note that these are all the same models as reported in Table \ref{tab:eng_perf}, except applied only to English instead of the target language.  

In all cases, the performance change is very small.  For micro-average there is a small performance increase, while there are small decreases in performance for \Fone and non-nil \Fone.  However, the largest drop in performance is less than $0.05$.  This illustrates the capacity of a model to adapt to a new language yet maintain a high-level of performance on the original training language.

\subsection{Design of Adversarial objective} \label{sec:adv_design}

How does the configuration of the \adv model change the performance of the model?  In Table \ref{tab:adv_params}, we explore how the performance on \tac evaluation changes when varying three factors -- the type of text, the size of the coefficient $\lambda$, and whether to train using the entity linking objective only for an additional 50 epochs.  For lower $\lambda$ and additional entity linking training, we found that both worked better on \wiki development data, while a higher $\lambda$ and full training worked better for \tac.  We also wanted to test the effect of using different text for the adversarial training.  Initially, we used randomly selected names in both English and the target language.  However, it is possible that the model better learns names instead of learning language-invariant representations.  Therefore, we also explore using the first $512$ subwords of randomly selected descriptions instead.

We report results in Table \ref{tab:adv_params}, with all non-baseline models trained on the joint entity linking and adversarial objective for 50 epochs, and the +EL models trained on EL data for an additional 50. Our reported setting for \tac, $\lambda = 0.25$ with name data, performs best on recall, \Fone, and non-NIL \Fone.  However, when using the description data and $\lambda = 0.01$ with or without additional EL training, a better micro-averaged precision is achieved.  Generally, the models using name data perform slightly better than those using descriptions, but the overall difference is slight (\eg \plus009 \Fone for $\lambda = 0.25$ with name, \minus015 \Fone with description).  Finally, recall generally performs best with a higher $\lambda$ and full adversarial training, and improves less with a lower $\lambda$ and EL only training.

\subsection{Analysis}

In general, \adv improves performance more on \tac data (Spanish and Chinese) than \wiki data (Farsi and Russian).  There are likely several factors behind this trend.  First, the distribution of mentions are different between the two datasets.  The similarity between mentions and entity names -- one measure of how easy the mentions are to link -- is much higher in \wiki.  For Farsi development mentions, 54.5\% were exact matches and had a Jaro-Winkler \cite{winkler1990string} lexical similarity of 94.1\%.  Compared to Spanish \tac (21.1\% exact, 71.4\% similarity) and Chinese (28\% exact, 66.1\% similarity), the Farsi data is relatively easy to link.

Second, when comparing the drop in performance from an in-language trained model to an English trained model, recall drops in the \tac data, while precision drops in the \wiki data.  The drop in precision may be due to the fact that we use English \tac data to train the zero-shot \wiki models, and that recall is fairly easy given the high mention-entity similarity.  Regardless, we see an increase in recall in the \wiki models, but this does not cancel out the reduction in precision.

\section{Related Work}

A long line of work on entity linking has developed standard models to link textual mentions to entities in a KB~\cite{dredze2010entity,durrett-klein-2014-joint,gupta-etal-2017-entity}. The models in this area have served as the basis for developing multilingual and cross-language entity linking systems, and they inform our own model development.  We define {\bf multilingual} to mean a model that can operate on mentions from more than one language at the same time (a single model to link both English mentions to an English KB and Chinese mentions to a Chinese KB) and {\bf cross-lingual} to refer to linking mentions in one language (\eg Chinese) to an ontology in another (\eg English).

% TODO Expand multilingual part, why not a baseline?
\citet{raiman2018deeptype} consider multilingual entity linking, in which they use a KB in the same language as the mention, but exploit multilingual transfer for the model's type system. They formulate a type system as a mixed integer problem, which they use to learn a type system from knowledge graph relations.

A larger body of work has explored the related task of cross-lingual entity linking.  Several approaches have been applied, including transliteration \cite{mcnamee-etal-2011-cross, pan-etal-2017-cross}, or translating the document into the language of the knowledge base and using a mono-lingual entity linking approach \cite{Ji2015}.  Additionally, many systems \cite{Tsai2016, upadhyay-etal-2018-joint}  leverage the cross-lingual structure of wikipedia to build entity linkers for that setting.  Newer work \citet{rijhwani2019zero} investigates zero-shot cross-lingual entity linking on low-resource languages.

\section{Conclusion}

In this work, we explore how to build a multi-lingual entity linking system that can effectively applied in zero-shot settings, where large amounts of annotated entity linking training data is unavailable.  With a neural ranker model using XLM-R, we see that while in-language trained models perform better than English-trained models applied to second languages, the performance decrease is not large.  

Next, we explore ways of improving these zero-shot models to reduce the performance loss compared to in-language models.  We specifically find that using an adversarial language classifier, which uses unannotated data, improves recall and \Fone on many datasets.  We also find that by adjusting the adversarial parameters, difference performance objectives can be achieved, such as maximizing recall or micro-averaged precision.  Overall, we find that training the model to learn language-invariant representations is effective in improving performance.

\bibliographystyle{acl_natbib}
\bibliography{anthology,multi}

\begin{thebibliography}{16}
\expandafter\ifx\csname natexlab\endcsname\relax\def\natexlab#1{#1}\fi

\bibitem[{Chen and Cardie(2018)}]{chen-cardie-2018-multinomial}
Xilun Chen and Claire Cardie. 2018.
\newblock \href {https://doi.org/10.18653/v1/N18-1111} {Multinomial adversarial
  networks for multi-domain text classification}.
\newblock In \emph{Proceedings of the 2018 Conference of the North {A}merican
  Chapter of the Association for Computational Linguistics: Human Language
  Technologies, Volume 1 (Long Papers)}, pages 1226--1240, New Orleans,
  Louisiana. Association for Computational Linguistics.

\bibitem[{Conneau et~al.(2019)Conneau, Khandelwal, Goyal, Chaudhary, Wenzek,
  Guzm{\'a}n, Grave, Ott, Zettlemoyer, and Stoyanov}]{conneau2019unsupervised}
Alexis Conneau, Kartikay Khandelwal, Naman Goyal, Vishrav Chaudhary, Guillaume
  Wenzek, Francisco Guzm{\'a}n, Edouard Grave, Myle Ott, Luke Zettlemoyer, and
  Veselin Stoyanov. 2019.
\newblock Unsupervised cross-lingual representation learning at scale.
\newblock \emph{arXiv preprint arXiv:1911.02116}.

\bibitem[{Dehghani et~al.(2017)Dehghani, Zamani, Severyn, Kamps, and
  Croft}]{dehghani2017neural}
Mostafa Dehghani, Hamed Zamani, Aliaksei Severyn, Jaap Kamps, and W~Bruce
  Croft. 2017.
\newblock Neural ranking models with weak supervision.
\newblock In \emph{Proceedings of the 40th International ACM SIGIR Conference
  on Research and Development in Information Retrieval}, pages 65--74. ACM.

\bibitem[{Devlin et~al.(2019)Devlin, Chang, Lee, and
  Toutanova}]{devlin-etal-2019-bert}
Jacob Devlin, Ming-Wei Chang, Kenton Lee, and Kristina Toutanova. 2019.
\newblock \href {https://doi.org/10.18653/v1/N19-1423} {{BERT}: Pre-training of
  deep bidirectional transformers for language understanding}.
\newblock In \emph{Proceedings of the 2019 Conference of the North {A}merican
  Chapter of the Association for Computational Linguistics: Human Language
  Technologies, Volume 1 (Long and Short Papers)}, pages 4171--4186,
  Minneapolis, Minnesota. Association for Computational Linguistics.

\bibitem[{Dredze et~al.(2010)Dredze, McNamee, Rao, Gerber, and
  Finin}]{dredze2010entity}
Mark Dredze, Paul McNamee, Delip Rao, Adam Gerber, and Tim Finin. 2010.
\newblock Entity disambiguation for knowledge base population.
\newblock In \emph{Conference on Computational Linguistics (COLING)}, pages
  277--285. Association for Computational Linguistics.

\bibitem[{Durrett and Klein(2014)}]{durrett-klein-2014-joint}
Greg Durrett and Dan Klein. 2014.
\newblock \href {https://doi.org/10.1162/tacl_a_00197} {A joint model for
  entity analysis: Coreference, typing, and linking}.
\newblock \emph{Transactions of the Association for Computational Linguistics},
  2:477--490.

\bibitem[{Gupta et~al.(2017)Gupta, Singh, and Roth}]{gupta-etal-2017-entity}
Nitish Gupta, Sameer Singh, and Dan Roth. 2017.
\newblock \href {https://doi.org/10.18653/v1/D17-1284} {Entity linking via
  joint encoding of types, descriptions, and context}.
\newblock In \emph{Proceedings of the 2017 Conference on Empirical Methods in
  Natural Language Processing}, pages 2681--2690, Copenhagen, Denmark.
  Association for Computational Linguistics.

\bibitem[{Ji et~al.(2015)Ji, Nothman, Hachey, and Florian}]{Ji2015}
Heng Ji, Joel Nothman, Ben Hachey, and Radu Florian. 2015.
\newblock \href
  {https://www.semanticscholar.org/paper/Overview-of-TAC-KBP2015-Tri-lingual-Entity-and-Ji-Nothman/955a78a8a5e4e31d10ffc827f365bd4c4f30d563}
  {{Overview of TAC-KBP2015 Tri-lingual Entity Discovery and Linking}}.
\newblock \emph{TAC}.

\bibitem[{McNamee et~al.(2011)McNamee, Mayfield, Lawrie, Oard, and
  Doermann}]{mcnamee-etal-2011-cross}
Paul McNamee, James Mayfield, Dawn Lawrie, Douglas Oard, and David Doermann.
  2011.
\newblock \href {https://www.aclweb.org/anthology/I11-1029} {Cross-language
  entity linking}.
\newblock In \emph{Proceedings of 5th International Joint Conference on Natural
  Language Processing}, pages 255--263, Chiang Mai, Thailand. Asian Federation
  of Natural Language Processing.

\bibitem[{Pan et~al.(2017{\natexlab{a}})Pan, Zhang, May, Nothman, Knight, and
  Ji}]{pan2017cross}
Xiaoman Pan, Boliang Zhang, Jonathan May, Joel Nothman, Kevin Knight, and Heng
  Ji. 2017{\natexlab{a}}.
\newblock Cross-lingual name tagging and linking for 282 languages.
\newblock In \emph{Proceedings of the 55th Annual Meeting of the Association
  for Computational Linguistics (Volume 1: Long Papers)}, volume~1, pages
  1946--1958.

\bibitem[{Pan et~al.(2017{\natexlab{b}})Pan, Zhang, May, Nothman, Knight, and
  Ji}]{pan-etal-2017-cross}
Xiaoman Pan, Boliang Zhang, Jonathan May, Joel Nothman, Kevin Knight, and Heng
  Ji. 2017{\natexlab{b}}.
\newblock \href {https://doi.org/10.18653/v1/P17-1178} {Cross-lingual name
  tagging and linking for 282 languages}.
\newblock In \emph{Proceedings of the 55th Annual Meeting of the Association
  for Computational Linguistics (Volume 1: Long Papers)}, pages 1946--1958,
  Vancouver, Canada. Association for Computational Linguistics.

\bibitem[{Raiman and Raiman(2018)}]{raiman2018deeptype}
Jonathan~Raphael Raiman and Olivier~Michel Raiman. 2018.
\newblock Deeptype: multilingual entity linking by neural type system
  evolution.
\newblock In \emph{Thirty-Second AAAI Conference on Artificial Intelligence}.

\bibitem[{Rijhwani et~al.(2019)Rijhwani, Xie, Neubig, and
  Carbonell}]{rijhwani2019zero}
Shruti Rijhwani, Jiateng Xie, Graham Neubig, and Jaime Carbonell. 2019.
\newblock Zero-shot neural transfer for cross-lingual entity linking.
\newblock In \emph{Proceedings of the AAAI Conference on Artificial
  Intelligence}, volume~33, pages 6924--6931.

\bibitem[{Tsai and Roth(2016)}]{Tsai2016}
Chen-Tse Tsai and Dan Roth. 2016.
\newblock \href {https://doi.org/10.18653/v1/N16-1072} {{Cross-lingual
  Wikification Using Multilingual Embeddings}}.
\newblock In \emph{Proceedings of the 2016 Conference of the North American
  Chapter of the Association for Computational Linguistics: Human Language
  Technologies}, pages 589--598, Stroudsburg, PA, USA. Association for
  Computational Linguistics.

\bibitem[{Upadhyay et~al.(2018)Upadhyay, Gupta, and
  Roth}]{upadhyay-etal-2018-joint}
Shyam Upadhyay, Nitish Gupta, and Dan Roth. 2018.
\newblock \href {https://doi.org/10.18653/v1/D18-1270} {Joint multilingual
  supervision for cross-lingual entity linking}.
\newblock In \emph{Proceedings of the 2018 Conference on Empirical Methods in
  Natural Language Processing}, pages 2486--2495, Brussels, Belgium.
  Association for Computational Linguistics.

\bibitem[{Winkler(1990)}]{winkler1990string}
William~E Winkler. 1990.
\newblock String comparator metrics and enhanced decision rules in the
  fellegi-sunter model of record linkage.
\newblock \emph{ERIC}.

\end{thebibliography}

\end{document}

\begin{appendix}
\section{Architecture information}\label{app:arch}

\begin{table}[h]
\centering
\begin{minipage}{0.8\textwidth}
\begin{tabular}{l|p{7cm}}
Parameter & Values \\ 
Context Layer(s) & [768], \textbf{[512]}, [256], [512,256] \\
Mention Layer(s) & [768], \textbf{[512]}, [256], [512,256] \\
Final Layer(s) & \textbf{[512,256]}, [256,128], [128,64], [1024,512], [512], [256] \\
Dropout probability & 0.1, \textbf{0.2}, 0.5 \\
Learning rate & 1e-5, 5e-4, \textbf{1e-4}, 5e-3, 1e-3 \\

\end{tabular} 
\renewcommand\thetable{6} 
\caption{To select parameters for the ranker, we tried 10 random combinations of the above parameters, and selected the configuration that performed best on the TAC development set.  The selected parameter is in bold. The full TAC multilingual model takes approximately 1 day to train on a single NVIDIA GeForce Titan RTX GPU, including candidate generation, representation caching, and prediction on the full evaluation dataset -- the Wiki model takes approximately 12 hours for the same set of steps.}
\end{minipage}
\end{table}

\section{TAC Triage Implementation}\label{app:triage}
We use the system discussed in  for both the \textbf{TAC} and \textbf{Wiki} datasets.  However, while the triage system provides candidates in the same KB as the \textbf{Wiki} data, not all entities in the \textbf{TAC} KB have Wikipedia page titles.  Therefore, the \textbf{TAC} triage step requires an intermediate step - using the Wikipedia titles generated by triage ($k=10$), we query a Lucene database of BaseKB for relevant entities.  For each title, we query BaseKB proportional to the prior provided by the triage system, meaning that we retrieve more BaseKB entities for titles that have a higher triage score, resulting in $l=200$ entities.  First, entities with Wikipedia titles are queried, followed by the entity name itself.  If none are found, we query the mention string - this provides a small increase in triage recall. 

\begin{table*}[]
    \centering
    \small
\begin{tabular}{ll|cccc|cccc|c}
& & \multicolumn{4}{c}{ All } & \multicolumn{4}{c}{ Non-NIL } & \\ 
Train/Test & Model & micro & p & r & f1 & micro & p & r & f1 & Eval Epoch \\ \hline
zh/zh & Baseline & 0.795 & 0.890 & 0.830 & 0.859 & 0.801 & 0.884 & 0.884 & 0.884 & 50 \\
en/zh & Baseline & 0.202 & 0.905 & 0.697 & 0.788 & 0.077 & 0.899 & 0.721 & 0.800 & 100 \\
en/zh & +A & 0.439 & 0.897 & 0.732 & 0.806 & 0.367 & 0.892 & 0.764 & 0.823 & 50 \\
en/zh & +PA & 0.635 & 0.889 & 0.753 & 0.815 & 0.606 & 0.881 & 0.789 & 0.833 & 100 \\
en/zh & +A (Desc) & 0.266 & 0.908 & 0.718 & 0.802 & 0.156 & 0.903 & 0.747 & 0.818 & \\
en/zh & +PA (Desc) & 0.645 & 0.885 & 0.774 & 0.826 & 0.618 & 0.877 & 0.815 & 0.845 & \\
en/zh & +P & 0.544 & 0.894 & 0.685 & 0.776 & 0.494 & 0.888 & 0.707 & 0.787 & 200 \\ \hline \hline
es/es & Baseline & 0.714 & 0.933 & 0.777 & 0.848 & 0.739 & 0.930 & 0.891 & 0.910 & 50 \\
en/es & Baseline & 0.488 & 0.942 & 0.643 & 0.764 & 0.444 & 0.944 & 0.716 & 0.815 & 100 \\
en/es & +A & 0.469 & 0.938 & 0.693 & 0.797 & 0.420 & 0.939 & 0.782 & 0.853 & 150 \\
en/es & +PA & 0.654 & 0.931 & 0.695 & 0.796 & 0.660 & 0.931 & 0.784 & 0.851 & 100 \\
en/es & +A (Desc) & 0.496 & 0.943 & 0.737 & 0.828 & 0.455 & 0.949 & 0.839 & 0.891 & \\
en/es & +PA (Desc) & 0.650 & 0.937 & 0.692 & 0.796 & 0.656 & 0.939 & 0.780 & 0.852 & \\
en/es & +P & 0.664 & 0.928 & 0.698 & 0.797 & 0.674 & 0.930 & 0.788 & 0.853 & 150 \\
\end{tabular} 
    \caption{Single runs of Development TAC results for our reported models, and the training epoch we report for that configuration in the evaluation results table.  Note that while we report results with the training sets equalized (zh and en training are set to be of equal size) for evaluation, the full development results do not have equalized training set sizes.}
    \label{tab:tac_dev}
\end{table*}

\begin{table*}[]
    \centering
    \small
    \begin{tabular}{cc|cccc|c}
Train/Test & Model & micro & p & r & f1 & Eval Epoch \\ \hline
ru/ru & Baseline & 0.650 & 0.823 & 0.888 & 0.854 & 800 \\
en/ru & Baseline & 0.484 & 0.762 & 0.855 & 0.806 & 550 \\
en/ru & +A & 0.451 & 0.712 & 0.893 & 0.792 & 50 \\
en/ru & +P & 0.473 & 0.685 & 0.860 & 0.762 & 50 \\ \hline \hline
fa/fa & Baseline & 0.832 & 0.881 & 0.966 & 0.922 & 800 \\
en/fa & Baseline & 0.603 & 0.720 & 0.928 & 0.811 & 150 \\
en/fa & +A & 0.447 & 0.555 & 0.948 & 0.700 & 200 \\
\end{tabular}

    \caption{Single runs of Development Wiki results for select reported models, and the training epoch we report for that configuration in the evaluation results table.  Note that while we report results with the training sets equalized (ru and en training are set to be of equal size) for evaluation, the full development results do not have equalized training set sizes.  For the +AP model, we report at Epoch 150 for Russian and 200 for Farsi, and for +P Farsi we report Epoch 50 (same as in Russian).  }
    \label{tab:wiki_dev}
\end{table*}

\begin{table*}[t]

    \centering
\begin{tabular}{ccc|ccc|ccc|ccc}
& & & \multicolumn{3}{c|}{In-Language} & \multicolumn{3}{c|}{En} & \multicolumn{3}{c}{En+A} \\
type & lang & count & micro & r & f1 & micro & r & f1 & micro & r & f1 \\ \hline
CMN & FAC & 59 & 0.169 & 0.631 & 0.756 & 0.119 & 0.515 & 0.670 & 0.169 & 0.632 & 0.768 \\
CMN & GPE & 3933 & 0.856 & 0.906 & 0.912 & 0.108 & 0.685 & 0.796 & 0.510 & 0.887 & 0.916 \\
CMN & LOC & 461 & 0.729 & 0.947 & 0.886 & 0.488 & 0.810 & 0.840 & 0.547 & 0.933 & 0.892 \\
CMN & ORG & 1441 & 0.160 & 0.726 & 0.774 & 0.299 & 0.629 & 0.722 & 0.127 & 0.799 & 0.821 \\
CMN & PER & 3116 & 0.708 & 0.682 & 0.797 & 0.612 & 0.676 & 0.792 & 0.610 & 0.676 & 0.792 \\ \hline
SPA & FAC & 59 & 0.051 & 0.294 & 0.454 & 0.068 & 0.285 & 0.444 & 0.102 & 0.289 & 0.448 \\
SPA & GPE & 1570 & 0.664 & 0.891 & 0.927 & 0.338 & 0.674 & 0.791 & 0.532 & 0.830 & 0.888 \\
SPA & LOC & 174 & 0.144 & 0.824 & 0.874 & 0.672 & 0.717 & 0.810 & 0.787 & 0.863 & 0.892 \\
SPA & ORG & 799 & 0.451 & 0.681 & 0.782 & 0.444 & 0.678 & 0.779 & 0.444 & 0.691 & 0.788 \\
SPA & PER & 2022 & 0.715 & 0.624 & 0.755 & 0.693 & 0.602 & 0.741 & 0.723 & 0.624 & 0.755 \\
\end{tabular}
\caption{How do the results of in-language training compare to English-only trained models and models trained with the adversarial objective?  We find that some types perform consistently, such as PER (or Persons) even in langauges that do not share scripts. Others, such as GPE (Geo-Political Entities) and ORG (Organizations) see a substantial drop in performance when applying a English-only model, but see more of that regained when using an adversarial objective.  These results are taken from a single run of the TAC evaluation data.}
\end{table*}
\end{appendix}